\documentclass[10pt,twocolumn,letterpaper]{article}

\usepackage{iccv}
\usepackage{times}
\usepackage{epsfig}
\usepackage{graphicx}
\usepackage{amsmath}
\usepackage{amssymb}
\usepackage{multirow}
\usepackage{booktabs}
\usepackage[font=normal]{subfig}
\usepackage{makecell}
\usepackage{bbm}
\usepackage{siunitx}
\usepackage{adjustbox}
\usepackage{tabu, booktabs}
\usepackage{makecell}
\usepackage{authblk}
\usepackage[dvipsnames]{xcolor}
\graphicspath{{figures/}}

\newcommand*{\affaddr}[1]{#1} 
\newcommand*{\affmark}[1][*]{\textsuperscript{#1}}

\makeatletter
\newcommand{\printfnsymbol}[1]{\textsuperscript{\@fnsymbol{#1}}}
\makeatother

\usepackage[breaklinks=true,bookmarks=false]{hyperref}

\iccvfinalcopy 

\def\iccvPaperID{****} 

\ificcvfinal\fi

\begin{document}

\title{ConvNets vs. Transformers: Whose Visual Representations are More Transferable?}

\author{%
	Hong-Yu Zhou\affmark[1]\quad Chixiang Lu\affmark[1]\quad Sibei Yang\affmark[2]\quad Yizhou Yu\affmark[1] \\
	\affaddr{\affmark[1]The University of Hong Kong}\quad
	\affaddr{\affmark[2]ShanghaiTech University}\\
	\tt\small{\{whuzhouhongyu, luchixiang\}@gmail.com, yangsb@shanghaitech.edu.cn, yizhouy@acm.org}\\
}


\maketitle
\ificcvfinal\thispagestyle{empty}\fi

\begin{abstract}
Vision transformers have attracted much attention from computer vision researchers as they are not restricted to the spatial inductive bias of ConvNets. However, although Transformer-based backbones have achieved much progress on ImageNet classification, it is still unclear whether the learned representations are as transferable as or even more transferable than ConvNets' features. To address this point, we systematically investigate the transfer learning ability of ConvNets and vision transformers in 15 single-task and multi-task performance evaluations. Given the strong correlation between the performance of pre-trained models and transfer learning, we include 2 residual ConvNets (i.e., R-101$\times$3 and R-152$\times$4) and 3 Transformer-based visual backbones (i.e., ViT-B, ViT-L and Swin-B), which have close error rates on ImageNet, that indicate similar transfer learning performance on downstream datasets. 

We observe consistent advantages of Transformer-based backbones on 13 downstream tasks (out of 15), including but not limited to fine-grained classification, scene recognition (classification, segmentation and depth estimation), open-domain classification, face recognition, etc. More specifically, we find that two ViT models heavily rely on whole network fine-tuning to achieve performance gains while Swin Transformer does not have such a requirement. Moreover, vision transformers behave more robustly in multi-task learning, i.e., bringing more improvements when managing mutually beneficial tasks and reducing performance losses when tackling irrelevant tasks. We hope our discoveries can facilitate the exploration and exploitation of vision transformers in the future.
\end{abstract}

\section{Introduction}

Ever since AlexNet~\cite{krizhevsky2012imagenet} was introduced for ImageNet classification~\cite{deng2009imagenet}, convolutional neural networks (i.e., ConvNets) have become the de-facto choice in computer vision related applications. Over the past decade, researchers have made great efforts to improve the performance of ConvNets, including but not restricted to increasing the network depth with small convolutional kernels~\cite{simonyan2014very} and residual connections~\cite{he2016deep}, embedding aggregated multi-branch architectures~\cite{szegedy2015going,xie2017aggregated} and automatically searching for neural architectures~\cite{zoph2016neural}. Nonetheless, the fundamental constraint of ConvNets, i.e., the inductive bias assumption towards local spatial structures, still remains, making ConvNets naturally disadvantageous in modeling long-range dependencies that are necessary for conducting logical reasoning.
\begin{figure}[t]
    \centering
    \includegraphics[width=0.8\columnwidth]{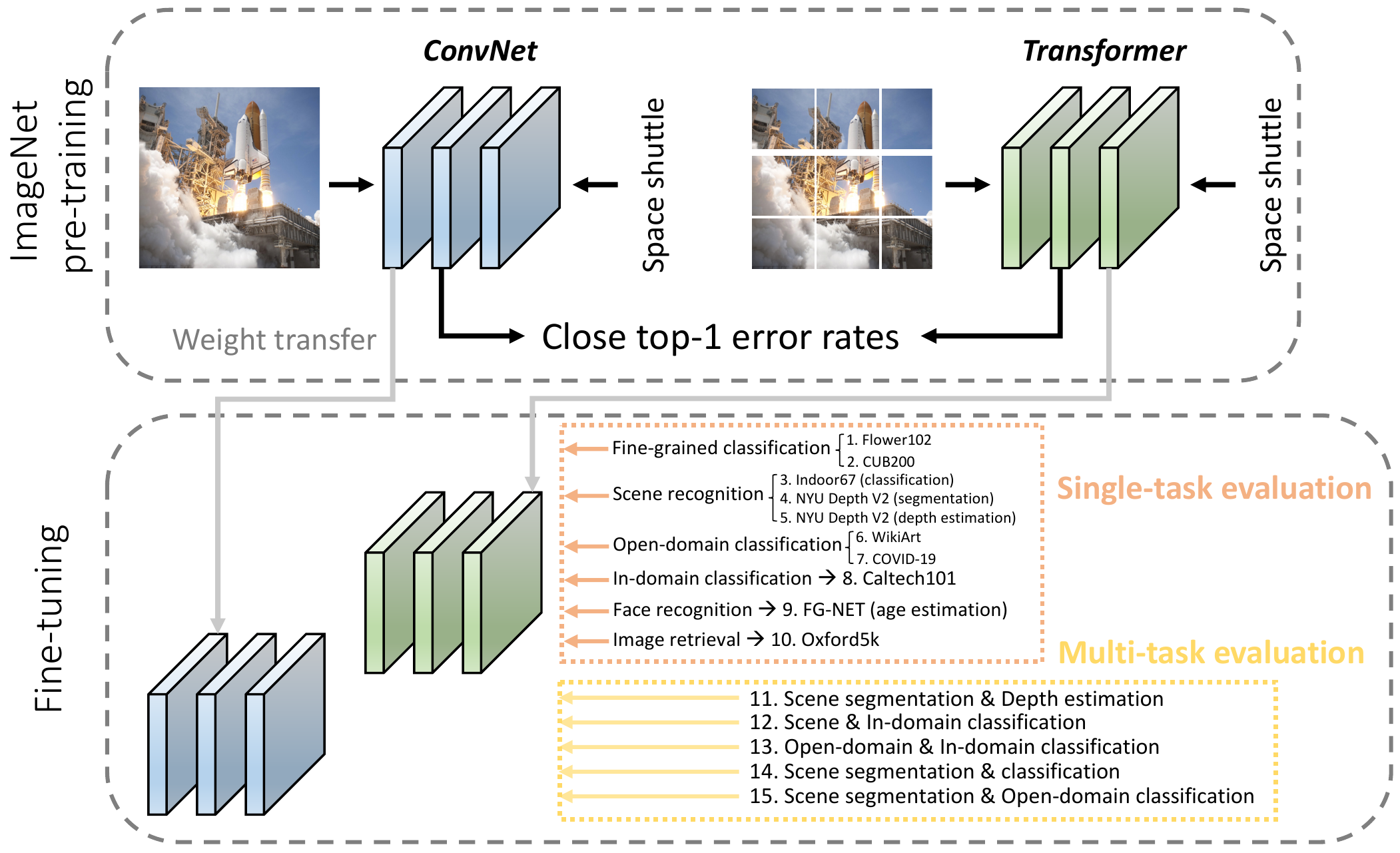}
    \caption{Overview of our investigation procedure. We ask pre-trained ConvNet and Transformer models to have close top-1 error rates on ImageNet classification. The pre-trained weights are then transferred to 15 downstream tasks (i.e., 10 single-task and 5 multi-task duties), to evaluate the transferability of learned representations.}
    \label{intro}
\end{figure}

On the other hand, inspired by the attention mechanism \cite{bahdanau2014neural}, Transformers~\cite{vaswani2017attention} remove convolutional and recurrent operations and solely rely on self-attention to model global dependencies between input and output. Meanwhile, compared to atypical recurrent models, Transformers greatly improve the training efficiency by allowing for large-scale parallelization. Based on above characteristics, Transformers have been the default model choice in various applications of natural language processing (NLP). In light of the successes that Transformers have achieved, many efforts have been made to extend Transformers to the computer vision field (i.e., building Vision Transformer), where pre-training a general-purpose Transformer-based visual backbone is one of the most promising directions and thus attracts much attention of the community. 
Vision Transformer (ViT)~\cite{dosovitskiy2020image} showed that Transformers can be extended to images to produce competitive ImageNet classification results in comparison to ResNet series \cite{he2016deep}. Recently, Liu \etal \cite{liu2021swin} proposed a hierarchical architecture, Swin Transformer, whose representation is computed with shifted windows. Swin Transformer reduces the quadratic computational complexity (with respect to image size) to linear, which promises higher training and inference efficiency. However, although Transformers have achieved results comparable to those of ConvNets on image classification, it is still unclear whether Transformers are able to provide equally transferable representations as ConvNets under the setting of transfer learning.

In this paper, we aim to investigate the transferability of the feature representations of both ConvNets and Transformers on a variety of downstream datasets following a schema of \emph{pre-training first, fine-tuning next}. Note that the scope of fine-tuning could be either the whole network or the last fully-connected layer (refer to linear evaluation protocol in Sec.~\ref{linear_eval}). Figure~\ref{intro} provides an overview of our investigation procedure. Specifically, we first pick two pre-trained ConvNet and Transformer based models, respectively, and require them to have similar top-1 error rates on ImageNet classification. The idea behind is that the accuracy of ImageNet-based pre-training has been shown to have a strong correlation with the accuracy of downstream fine-tuning~\cite{kornblith2019better}. Close top-1 errors indicate that the two pre-trained models should presumably have comparable transfer learning performance. Otherwise, if two pre-trained models have quite different performance on ImageNet, the comparison of their fine-tuning performance would be unfair and meaningless as they ought to have some performance differences in transfer learning. Next, we further optimize the pre-trained weights in the fine-tuning stage and evaluate the improved representations on 15 downstream tasks. Different from \cite{yosinski2014transferable,sharif2014cnn,sharif2014cnn} that evaluate ConvNet's features on single tasks, we conduct both single-task and multi-task learning for more extensive evaluation. The chosen downstream tasks cover a variety of recognition problems, including fine-grained classification (Flower102 and CUB200), indoor scene classification (Indoor67), scene segmentation and depth estimation (NYU Depth V2), in-domain (Caltech101) and open-domain classification (WikiArt and COVID-19).

There are three aspects in our findings. First, transformer-based backbones are more advantageous than ConvNets when transfer learning is performed on downstream data that have large domain gaps with ImageNet, including but not restricted to fine-grained classification, scene recognition (i.e., classification, segmentation and depth estimation), open-domain classification and face recognition. We believe the above observation provides a strong evidence that Transformer-based backbones produce more generalizable and transferable representations than ConvNet-based models. Second, we observe that the performance advantages (over ConvNets) of two ViT backbones are largely due to whole network fine-tuning, whereas Swin-B does not have such a requirement. Last but not the least, it appears that vision transformers are more robust in multi-task evaluation. More specifically, Transformer-based backbones bring larger improvements when multiple tasks are complementary, while producing smaller performance drops when tasks cannot benefit each other. We believe these advantages can be attributed to two strengths of vision transformers. First of all, they are naturally not confined to the local inductive bias of ConvNets and thus have the ability to capture long-range dependencies. Second, Transformer-based backbones often have much fewer network parameters compared to ConvNets with similar pre-training performance on ImageNet, which would reduce the risk of overfitting when they are transferred to small-scale downstream datasets. When comparing Swin Transformer with two ViT models, we believe the pyramidal feature hierarchy in Swin-B produces more transferable visual features during the pre-training stage and reduces its dependency on whole network fine-tuning.


\begin{table}[t]
    \centering
    \scriptsize
    \begin{tabular}{c|c|c|c|c}
        \toprule
         Type & Model & IN (acc.) $\uparrow$ & Params. & Ave. rank $\downarrow$ \\
         \hline
         \hline
         \multirow{2}{*}{ConvNet} & R-101$\times$3 & 84.4 ({\color{gray}4}) & 388M & \multirow{2}{*}{{\color{gray}2.5}}\\
         & R-152$\times$4 & 85.4 ({\color{gray}1}) & 937M & \\
         \hline
         \multirow{3}{*}{Transformer} & ViT-B/16 & 84.0 ({\color{gray}5}) & 86M & \multirow{3}{*}{{\color{gray}3.0}}\\
         & ViT-L/16 & 85.2 ({\color{gray}2}) & 307M\\
         & Swin-B & 85.2 ({\color{gray}2}) & 88M\\
         \bottomrule
    \end{tabular}
    \caption{Involved pre-trained models. \textbf{/16} denotes the 16×16 input patch size. All models are pre-trained on ImageNet-21k and tested on ImageNet-1k. \textbf{IN} is an abbreviation for ImageNet. We display the top-1 accuracy and corresponding performance rank (in a descending order) on Image-1k. For two groups of models (i.e., ConvNet-based and Transformer-based), we present their average ranks, respectively. $\uparrow$ denotes the higher the better while $\downarrow$ stands for the opposite.}
    \label{in}
\end{table}
\section{Pre-trained models to be evaluated}
There are 5 different pre-trained models that serve as backbones in our experiments, as displayed in Table \ref{in}. For ConvNet-based backbones, we choose to use deep residual networks \cite{he2016deep} that are among the most effective deep neural networks with hand-crafted architectures. R-101$\times$3 and R-152$\times$4 are 3$\times$ and 4$\times$ wider ResNet-101 and ResNet-152, respectively, which are pre-trained with carefully selected strategies \cite{kolesnikov2020big}. For Transformer-based backbones, we pick ViT-B/16, ViT-L/16 and Swin-B, which are all representative vision transformer backbones. ViT-B/16 and ViT-L/16 comprises alternating layers of multi-head self-attention (MSA) and multi-layer perceptron (MLP) blocks:
\begin{align}
\begin{split}
    \textbf{\text{z}}_{l+1}^{\prime} &= \text{MSA}(\text{LN}(\textbf{\text{z}}_{l})) + \textbf{\text{z}}_{l},\\
    \textbf{\text{z}}_{l+1} &= \text{MLP}(\text{LN}(\textbf{\text{z}}_{l+1}^{\prime})) + \textbf{\text{z}}_{l+1}^{\prime},
\end{split}
\end{align}
where $l$ denotes the layer index and $\textbf{\text{z}}_l$ stands for the input of layer. Swin Transformer improves ViT by replacing layers of MSA and MLP with window-based and shifted-window-based multi-head self-attention (i.e., W-MSA and SW-MSA) with two MLP blocks, which can dramatically reduce the computational complexity:
\begin{align}
\begin{split}
    \Tilde{\textbf{\text{z}}}_{l+1}^{\prime} &= \text{W-MSA}(\text{LN}(\textbf{\text{z}}_{l})) + \textbf{\text{z}}_{l},\\
    \textbf{\text{z}}_{l+1} &= \text{MLP}(\text{LN}(\Tilde{\textbf{\text{z}}}_{l+1}^{\prime})) + \Tilde{\textbf{\text{z}}}_{l+1}^{\prime},\\
    \Tilde{\textbf{\text{z}}}_{l+2}^{\prime} &= \text{SW-MSA}(\text{LN}(\textbf{\text{z}}_{l+1})) + \textbf{\text{z}}_{l+1},\\
    \textbf{\text{z}}_{l+2} &= \text{MLP}(\text{LN}(\Tilde{\textbf{\text{z}}}_{l+2}^{\prime})) + \Tilde{\textbf{\text{z}}}_{l+2}^{\prime}.\\
\end{split}
\end{align}

From Table \ref{in}, we can see that the average performance rank (on ImageNet-1k) of ConvNet-based models is higher than that of Transformer-based. According to the comprehensive investigation in \cite{kornblith2019better}, ConvNet-based models would probably achieve better (at least comparable) transfer learning results than Transformer-based networks. 

\section{Single-task evaluation}
We include 10 tasks in single-task evaluation, which consists of a range of topics, such as fine-grained classification, scene recognition, in-domain and open-domain classification, etc. The goal is to extensively evaluate the transferring ability of representations of both ConvNet and Transformer. In the following, we go through different tasks one-by-one. For each experiment, we repeat it for 3 times and report the average results.

\begin{table}[]
    \centering
    \scriptsize
    \begin{tabular}{c|c|c|c}
        \toprule
         Model & \scriptsize Flower102 (acc.) $\uparrow$ & \scriptsize CUB200 (acc.) $\uparrow$ & Ave. rank $\downarrow$ \\
         \hline
         \hline
         R-101$\times$3 & 98.4 ({\color{gray}{5}}) & 87.5 ({\color{gray}{5}}) & \multirow{2}{*}{{\color{gray}{4.3}}} \\
         R-152$\times$4 & 98.9 ({\color{gray}{4}}) & 88.6 ({\color{gray}{3}}) &  \\
         \hline
         ViT-B/16 & 99.2 ({\color{gray}{3}}) & 88.3 ({\color{gray}{4}}) & \multirow{3}{*}{{\color{gray}{2.2}}}\\
         ViT-L/16 & 99.4 ({\color{gray}{2}}) & 88.9 ({\color{gray}{2}})\\
         Swin-B & 99.8 ({\color{gray}{1}}) & 89.9 ({\color{gray}{1}})\\
         \bottomrule
    \end{tabular}
    \caption{Results on fine-grained classification. We mark the performance rank (in a descending order) using color gray. $\uparrow$ denotes the higher the better while $\downarrow$ stands for the opposite.}
    \label{fine-grained}
\end{table}
\subsection{Fine-grained classification}
\noindent \textbf{Flower102 \cite{sharif2014cnn}.} This dataset consists of 102 flower categories that are commonly occurring in UK. Each category contains 40 to 258 images. There are two main challenges of Flower102: i) large similarity between classes and ii) large variation within classes. \\

\noindent \textbf{CUB200 \cite{wah2011caltech}.} 200 bird species and 11,788 images are included in this dataset. The names of species were obtained using an online bird species guide and organized by scientific classification (order, family, genus, species). Flicker Image Search engine is used to acquire bird pictures, which are then filtered by human annotators. There are two versions of CUB200 \cite{welinder2010caltech,wah2011caltech} and we use the latest version \cite{wah2011caltech}.\\

\noindent \textbf{Implementation details.} Here we present training and testing strategies of fine-grained classification. Other tasks in this paper may share similar hyper-parameters, where we will clarify the differences of implementation. We conducted all experiments using PyTorch \cite{paszke2019pytorch}.
\begin{itemize}
    \item Data split: For Flower102, we randomly select 80\% images as the training set. The rest 20\% data are evenly divided into validation and test sets. For CUB200, we directly use the official test set. 10\% images are randomly selected from the official training set to build the validation set, while the rest training images form the training set.
    \item Network architecture: We append a classification head after each backbone, before which we add a dropout layer ($p$=0.2).
    \item Optimizer: Adam is used as the default optimizer, where $\beta_1$ is set to 0.9 and $\beta_2$ is set to 0.999. We set weight decay to 1e-6.
    \item Learning rate: The initial learning rate is 1e-4 and is decayed by a factor of 2 each time the validation loss stops decreasing after 3 epochs.
    \item Augmentation strategies: We use random crop, random rotation (-30 degrees to 30 degrees) and random horizontal flip. The input image size is 224$\times$224.
    \item Batch size and training time: The training batch size is 32. We stop the training procedure when the validation loss stops decreasing for up to 10 epochs.
    \item Loss function: We directly use the cross entropy loss.
    \item Other techniques: We use the label smoothing strategy, where the smoothing rate is set to 0.1. During the fine-tuning stage, we first freeze the whole backbone and conduct warm-up for 200 training iterations using a small learning rate (1e-6). Then, we fine-tune the whole network using the initial learning rate (i.e., 1e-4).
\end{itemize}

\noindent \textbf{Results.} We present the experimental results in Table \ref{fine-grained}. Somewhat surprisingly, we discover that Transformer-based backbones hold observable advantages over ConvNets, where the average rank of all 3 vision transformers is much higher than that of 2 ConvNet-based backbones. In contrast with the higher average rank that ConvNets have achieved on ImageNet classification (refer to Table \ref{in}), these outstanding results on Flower102 and CUB200 reflect the great discriminative and transferring abilities of transformer-based representations in capturing small differences. More specifically, Swin-B and ViT-L/16 are two best performing backbones on both datasets. ViT-B/16 that achieves the lowest top-1 accuracy on ImageNet-1k produces comparable results with R-152$\times$4 that maintains the highest top-1 accuracy in Table \ref{in}. These phenomena again verify the advantages of Transformer when dealing with fine-grained classification problems.

\subsection{Scene recognition}
\begin{table*}[t]
    \centering
    \scriptsize
    \begin{tabular}{c|c|c|c|c}
        \toprule
         Model & Indoor67 (acc.) $\uparrow$ & NYU segmentation (mIoU) $\uparrow$ & NYU depth estimation (RMSE) $\downarrow$ & Ave. rank $\downarrow$ \\
         \hline
         \hline
         R-101$\times$3 & 84.3 ({\color{gray}{5}}) & 52.2 ({\color{gray}{5}}) & 0.387 ({\color{gray}{5}})& \multirow{2}{*}{{\color{gray}{4.2}}} \\
         R-152$\times$4 & 85.7 ({\color{gray}{2}}) & 53.1 ({\color{gray}{4}}) & 0.382 ({\color{gray}{4}})& \\
         \hline
         ViT-B/16 & 85.1 ({\color{gray}{4}}) & 53.2 ({\color{gray}{3}}) & 0.368 ({\color{gray}{3}}) & \multirow{3}{*}{{\color{gray}{2.2}}}\\
         ViT-L/16 & 85.5 ({\color{gray}{3}}) & 53.4 ({\color{gray}{2}}) & 0.360 ({\color{gray}{2}}) & \\
         Swin-B & 87.6 ({\color{gray}{1}}) & 54.1 ({\color{gray}{1}}) & 0.358 ({\color{gray}{1}}) &\\
         \bottomrule
    \end{tabular}
    \caption{Results on scene recognition. mIoU and RMSE (Root Mean Square Error) are used as the evaluation metrics for segmentation and depth estimation tasks, respectively. We mark the performance rank (in a descending order) using color gray.}
    \label{scene}
\end{table*}
\noindent \textbf{Indoor67 \cite{quattoni2009recognizing}.} This database contains 67 indoor scene categories, and a total of 15,620 images. The number of images varies across categories, but each category includes at least 100 pictures. All images are collected using online image search engines and some of them come from LabelMe dataset. A minimum resolution of 200 pixels in the smallest axis is guaranteed.\\

\noindent \textbf{NYU Depth V2 \cite{silberman2012indoor}.} The dataset comprises rich annotations of both semantic segmentation and depth estimation for 35,064 distinct objects that are collected from 3 different US cities using Kinect. All 1,449 images are divided into 40 scene classes and a dense per-pixel labeling is conducted.\\

\noindent \textbf{Implementation details.} Most details of making experiments on Indoor67 are similar to those of Flower102 and CUB200. In the following, we mainly introduce how to implement indoor semantic segmentation and depth estimation.
\begin{itemize}
    \item Data split: For Indoor67, we use the official test list while randomly split 10\% data from the training list to build the validation set. The rest 90\% training data form the training set. For NYU Depth V2 (i.e., both segmentation and depth estimation tasks), we randomly select 80\% images as the training set. The validation and test sets include 10\% data, respectively.
    \item Network architecture: For indoor scene segmentation, we directly make use of the segmentation head of UPerNet \cite{xiao2018unified} and append it after ConvNet or Transformer backbones. As for depth estimation, we modify the segmentation head to predict depth values.
    \item Optimizer: We use AdamW \cite{loshchilov2017decoupled}, where $\beta_1$ is set to 0.9, $\beta_2$ is set to 0.999 and weight decay is set to 1e-6.
    \item Learning rate: For both tasks, the initial learning rate is 6e-5 and we employ a polynomial learning rate decay strategy where the power value is set to 0.9.
    \item Augmentation strategies: For semantic segmentation, we use random crop and random horizontal flip. We also apply a variety of photometric distortion, including but not limited to random brightness, random contrast, random saturation, etc. The input image size is 512$\times$512. For depth estimation, only random crop and random horizontal flip are applied, and the input size 384$\times$384.
    \item Batch size and training time: The training batch size is 8. The training procedure for each model lasts for 20,000 iterations.
    \item Loss function: For semantic segmentation, we apply the cross entropy loss (weight=1) and deep supervision loss (weight=0.4). For depth estimation, we employ the scale-and shift-invariant trimmed loss (weight=1) that operates on an inverse depth representation \cite{ranftl2021vision} and gradient-matching loss \cite{li2018megadepth} (weight=1).
    \item Other techniques: We conduct warm-up for 1,500 training iterations using a small learning rate (1e-6). Then, we fine-tune the whole network using the initial learning rate (i.e., 6e-5).\\
\end{itemize}

\noindent \textbf{Results.} Table \ref{scene} displays the experimental results of 3 tasks on scene recognition. Again, Transformer-based backbones prominently outperform ConvNet-based models in the average rank. Moreover, all 3 Transformer-based backbones occupy the top 3 positions on both indoor scene segmentation and depth estimation tasks. Considering recognizing scenes is a quite challenging task that requires strong reasoning ability to understand the relationship between objects and scenes, the obvious improvements brought by vision transformers demonstrate their advantages of mastering complex situations. Besides, we can also observe that Swin-B achieve the best performance on all 3 scene recognition tasks, suggesting the potential of producing high-quality transferable representations using efficient Transformer-based backbones.

\subsection{Open-domain classification}
\begin{table}[]
    \centering
    \scriptsize
    \begin{tabular}{c|c|c|c}
        \toprule
         Model & \scriptsize WikiArt (acc.) $\uparrow$ & \scriptsize COVID-19 (acc.) $\uparrow$ & Ave. rank $\downarrow$ \\
         \hline
         \hline
         R-101$\times$3 & 66.2 ({\color{gray}{5}}) & 80.6 ({\color{gray}{5}}) & \multirow{2}{*}{{\color{gray}{4.5}}} \\
         R-152$\times$4 & 66.4 ({\color{gray}{4}}) & 81.2 ({\color{gray}{4}}) &  \\
         \hline
         ViT-B/16 & 67.4 ({\color{gray}{3}}) & 81.7 ({\color{gray}{3}}) & \multirow{3}{*}{{\color{gray}{2.0}}}\\
         ViT-L/16 & 68.4 ({\color{gray}{2}}) & 82.1 ({\color{gray}{2}})\\
         Swin-B & 71.0 ({\color{gray}{1}}) & 82.6 ({\color{gray}{1}})\\
         \bottomrule
    \end{tabular}
    \caption{Results on open-domain classification. We mark the performance rank (the smaller the better) using color gray.}
    \label{open}
\end{table}
\noindent \textbf{WikiArt \cite{tan2016ceci}.} This database has a collection of more than 80,000 fine-art paintings from more than 1,000 artists, ranging from the 15-th century to modern times. All images are collected from \url {https://www.wikiart.org/} and can be divided into 27 different styles.\\

\noindent \textbf{COVID-19 Image Data Collection \cite{cohen2020covid}.} More than 700 pneumonia cases with chest X-rays are involved, which were built to improve the identification of COVID-19. These X-rays come from over 400 people from 26 countries. In this dataset, we mainly focus on distinguishing COVID-19 from other disease/normal images.\\

\noindent \textbf{Implementation details.} We implement classification networks on WikiArt and COVID-19 Image Data Collection using the same training strategies applied to Flower102 and CUB200. Specifically, in the task of COVID-19 classification, we build the training set using images from Australia and America. X-rays from Africa form the validation set while the remaining images are included in the test set.\\

\noindent \textbf{Results.} The results on open-domain classification are shown in Table \ref{open}, from which we can find similar phenomena that have been observed on fine-grained classification and scene recognition. Again and again, Transformer-based backbones surpass ConvNet-based models by large margins. More importantly, we can see that all 3 vision transformers occupy the top 3 places on both WikiArt and COVID-19 as what they have done on tasks of indoor scene segmentation and depth estimation. Considering the images of WikiArt and COVID-19 are not included in ImageNet, the improvements offered by Transformer-based representations provide strong evidences for their great transferability. In addition, we can see that Swin-B again becomes the winner on open-domain classification, verifying its representations are more transferable and generalizable than those of ViT-L/16 (with the same top-1 accuracy on ImageNet-1k classification).

\subsection{In-domain classification}
\begin{table}[]
    \centering
    \scriptsize
    \begin{tabular}{c|c|c}
        \toprule
         Model & Caltech101 (acc.) $\uparrow$ & Ave. rank \\
         \hline
         \hline
         R-101$\times$3 & 96.8 ({\color{gray}{2}}) & \multirow{2}{*}{{\color{gray}{2.5}}}  \\
         R-152$\times$4 & 96.7 ({\color{gray}{3}}) \\
         \hline
         ViT-B/16 & 96.7 ({\color{gray}{3}}) & \multirow{3}{*}{{\color{gray}{3.0}}}\\
         ViT-L/16 & 96.5 ({\color{gray}{5}})\\
         Swin-B & 97.7 ({\color{gray}{1}})\\
         \bottomrule
    \end{tabular}
    \caption{Results on in-domain classification. We mark the performance rank (the smaller the better) using color gray.}
    \label{caltech}
\end{table}
\noindent \textbf{Caltech101 \cite{fei2004learning}.} This dataset consists of 9,146 images, which are acquired by searching names of 101 categories using Google Image Search engine and filtering out irrelevant search results. Each category contains 40 to 800 pictures and most of them have about 50 images. The size of each image is roughly 300 x 200 pixels.\\

\noindent \textbf{Implementation details.} We simply employ the same training strategies on Flower102 and CUB200.\\

\noindent \textbf{Results.} It is not surprising to find ConvNet-based backbones achieve a higher average rank on in-domain classification as some classes and images of Caltech101 overlap with those in ImageNet. If we compare the average ranks in Table \ref{caltech} with those in Table \ref{in}, it is obvious that the they are similar and quite consistent. In other words, we can draw a conclusion that better ImageNet performance often lead to better results on Caltech101 as the backbones have the ability to memorize seen images and recognize similar classes.

\begin{table}[]
    \centering
    \scriptsize
    \begin{tabular}{c|c|c}
        \toprule
         Model & FG-NET (MAE) $\downarrow$ & Ave. rank $\downarrow$ \\
         \hline
         \hline
         R-101$\times$3 & 3.6 ({\color{gray}{3}}) & \multirow{2}{*}{{\color{gray}{4.0}}}  \\
         R-152$\times$4 & 4.7 ({\color{gray}{5}}) \\
         \hline
         ViT-B/16 & 3.5 ({\color{gray}{2}}) & \multirow{3}{*}{{\color{gray}{2.3}}}\\
         ViT-L/16 & 4.5 ({\color{gray}{4}})\\
         Swin-B & 3.0 ({\color{gray}{1}})\\
         \bottomrule
    \end{tabular}
    \caption{Results on face recognition (i.e., facial age estimation). MAE is defined as the average of the absolute errors between the estimated ages and the ground truth ages. We mark the performance rank (the smaller the better) using color gray.}
    \label{FG-NET}
\end{table}
\begin{table}[]
    \centering
    \scriptsize
    \begin{tabular}{c|c|c|c|c}
        \toprule
         Model & CUB200 $\uparrow$ & Indoor67 $\uparrow$ & WikiArt $\uparrow$ & Ave. rank $\downarrow$ \\
         \hline
         R-101$\times$3 & 84.7 ({\color{gray}{2}}) & 83.7 ({\color{gray}{4}}) & 53.3 ({\color{gray}{3}}) & \multirow{2}{*}{{\color{gray}{3.0}}} \\
         R-152$\times$4 & 75.2 ({\color{gray}{5}}) & 85.6 ({\color{gray}{2}}) & 53.9 ({\color{gray}{2}}) & \\
         \hline
         ViT-B/16 & 79.5 ({\color{gray}{4}}) & 81.8 ({\color{gray}{5}}) & 47.3 ({\color{gray}{4}}) & \multirow{3}{*}{{\color{gray}{3.0}}}\\
         ViT-L/16 & 80.1 ({\color{gray}{3}}) & 84.1 ({\color{gray}{3}}) & 46.4 ({\color{gray}{5}}) &\\
         Swin-B & 87.1 ({\color{gray}{1}}) & 86.3 ({\color{gray}{1}}) & 54.2 ({\color{gray}{1}}) &\\
         \bottomrule
    \end{tabular}
    \caption{Linear classification protocol. The evaluation metric is mean accuracy. We mark the performance rank (the smaller the better) using color gray.}
    \label{linear_cls}
\end{table}
\begin{table}[]
    \centering
    \scriptsize
    \begin{tabular}{c|c|c}
        \toprule
         Model & Oxford5k (mAP) $\uparrow$ & Ave. rank $\downarrow$ \\
         \hline
         \hline
         R-101$\times$3 & 60.7 ({\color{gray}{1}}) & \multirow{2}{*}{{\color{gray}{2.0}}}  \\
         R-152$\times$4 & 59.6 ({\color{gray}{3}}) \\
         \hline
         ViT-B/16 & 58.2 ({\color{gray}{4}}) & \multirow{3}{*}{{\color{gray}{3.7}}}\\
         ViT-L/16 & 57.7 ({\color{gray}{5}})\\
         Swin-B & 59.9 ({\color{gray}{2}})\\
         \bottomrule
    \end{tabular}
    \caption{Results on unsupervised image retrieval. mAP stands for mean average precision. We mark the performance rank (the smaller the better) using color gray.}
    \label{retrieval}
\end{table}
\begin{table*}[]
    \centering
    \scriptsize
    \begin{tabular}{c|c|c|c|c|c}
        \toprule
         \multirow{2}{*}{Model} & \multicolumn{2}{c|}{NYU segmentation (mIoU)} & \multicolumn{2}{c|}{NYU depth estimation (RMSE)} & \multirow{2}{*}{Ave. rank $\downarrow$} \\ \cline{2-5}
         & Perf. $\uparrow$ & Ave. {\color{ForestGreen}Imp. $\uparrow$} / {\color{red}Drop $\downarrow$} & Perf. $\downarrow$ & Ave. {\color{ForestGreen}Imp. $\uparrow$} / {\color{red}Drop $\downarrow$} &\\
         \hline
         R-101$\times$3 & 52.4 ({\color{gray}{5}}) & \multirow{2}{*}{\color{ForestGreen}{0.3}} & 0.381 ({\color{gray}{5}}) & \multirow{2}{*}{\color{ForestGreen}{0.005}} & \multirow{2}{*}{{\color{gray}{4.5}}} \\
         R-152$\times$4 & 53.5 ({\color{gray}{4}}) & & 0.379 ({\color{gray}{4}}) & & \\
         \hline
         ViT-B/16 & 53.9 ({\color{gray}{3}}) & \multirow{3}{*}{\color{ForestGreen}{0.4}} & 0.359 ({\color{gray}{3}}) & \multirow{3}{*}{\color{ForestGreen}{0.007}} & \multirow{3}{*}{{\color{gray}{2.0}}}\\
         ViT-L/16 & 53.8 ({\color{gray}{2}}) & & 0.352 ({\color{gray}{2}}) & \\
         Swin-B & 54.5 ({\color{gray}{1}}) & & 0.355 ({\color{gray}{1}}) & \\
         \bottomrule
    \end{tabular}
    \caption{Multi-task learning on scene segmentation and depth estimation. \textbf{Perf.} and \textbf{Imp.} are abbreviations for performance and improvement, respectively. The green/red color denotes the relative improvement/drop in performance using multi-task learning over using single datasets. The performance ranks (in a descending order) are marked using color grey.}
    \vspace{-1.0em}
    \label{scene_seg_depth}
\end{table*}
\begin{table*}[]
    \centering
    \scriptsize
    \begin{tabular}{c|c|c|c|c|c}
        \toprule
         \multirow{2}{*}{Model} & \multicolumn{2}{c|}{Indoor67 (acc.)} & \multicolumn{2}{c|}{Caltech101 (acc.)} & \multirow{2}{*}{Ave. rank $\downarrow$} \\ \cline{2-5}
         & Perf. $\uparrow$ & Ave. {\color{ForestGreen}Imp. $\uparrow$} / {\color{red}Drop $\downarrow$} & Perf. $\uparrow$ & Ave. {\color{ForestGreen}Imp. $\uparrow$} / {\color{red}Drop $\downarrow$} &\\
         \hline
         R-101$\times$3 & 82.4 ({\color{gray}{5}}) & \multirow{2}{*}{\color{red}{2.2}}& 95.9 ({\color{gray}{5}}) & \multirow{2}{*}{\color{red}{0.8}}& \multirow{2}{*}{{\color{gray}{4.3}}} \\
         R-152$\times$4 & 83.2 ({\color{gray}{3}}) & & 96.0 ({\color{gray}{4}}) &  \\
         \hline
         ViT-B/16 & 83.7 ({\color{gray}{2}}) & \multirow{3}{*}{\color{red}{1.8}}& 96.9 ({\color{gray}{3}}) &\multirow{3}{*}{\color{ForestGreen}{0.1}}& \multirow{3}{*}{{\color{gray}{3.3}}}\\
         ViT-L/16 & 82.9 ({\color{gray}{4}}) & & 97.0 ({\color{gray}{2}}) & &\\
         Swin-B & 86.3 ({\color{gray}{1}}) & & 97.4 ({\color{gray}{1}}) & &\\
         \bottomrule
    \end{tabular}
    \caption{Multi-task classification on scenes and generic objects. The green/red color denotes the relative improvement/drop in performance using multi-task learning over using single datasets. We mark the performance rank (the smaller the better) using color gray.}
    \vspace{-1.0em}
    \label{scene_and_in_domain}
\end{table*}
\begin{table*}[]
    \centering
    \scriptsize
    \begin{tabular}{c|c|c|c|c|c}
        \toprule
         Model & \multicolumn{2}{c|}{Caltech101 (acc.)} & \multicolumn{2}{c|}{WikiArt (acc.)} & Ave. rank $\downarrow$ \\
        \cline{2-5}
         & Perf. $\uparrow$ & Ave. {\color{ForestGreen}Imp. $\uparrow$} / {\color{red}Drop $\downarrow$} & Perf. $\uparrow$ & Ave. {\color{ForestGreen}Imp. $\uparrow$} / {\color{red}Drop $\downarrow$} & \\
         \hline
         R-101$\times$3 & 90.6 ({\color{gray}{4}}) & \multirow{2}{*}{\color{red}{6.7}}& 66.1 ({\color{gray}{5}}) & \multirow{2}{*}{\color{red}{0.1}} & \multirow{2}{*}{{\color{gray}{4.5}}} \\
         R-152$\times$4 & 89.6 ({\color{gray}{5}}) & & 66.3 ({\color{gray}{4}}) & & \\
         \hline
         ViT-B/16 & 95.6 ({\color{gray}{2}}) & \multirow{3}{*}{\color{red}{2.1}} &67.6 ({\color{gray}{3}}) & \multirow{3}{*}{\color{ForestGreen}{0.1}}& \multirow{3}{*}{{\color{gray}{2.0}}}\\
         ViT-L/16 & 92.9 ({\color{gray}{3}}) & & 67.8 ({\color{gray}{2}})&&\\
         Swin-B & 96.1 ({\color{gray}{1}}) & & 71.6 ({\color{gray}{1}})&&\\
         \bottomrule
    \end{tabular}
    \caption{Multi-task classification on art styles and generic objects. The performance ranks (in a descending order) are marked using color grey, while color green/red denotes the relative improvement/drop over single-task training.}
    \vspace{-1.0em}
    \label{open_and_in}
\end{table*}
\begin{table*}[]
    \centering
    \scriptsize
    \begin{tabular}{c|c|c|c|c|c}
        \toprule
         \multirow{2}{*}{Model} &  \multicolumn{2}{c|}{NYU segmentation (mIoU)} & \multicolumn{2}{c|}{Indoor67 (acc.)} & Ave. rank $\downarrow$ \\ \cline{2-5}
         & Perf. $\uparrow$ & Ave. {\color{ForestGreen}Imp. $\uparrow$} / {\color{red}Drop $\downarrow$} & Perf. $\uparrow$ & Ave. {\color{ForestGreen}Imp. $\uparrow$} / {\color{red}Drop $\downarrow$} &\\
         \hline
         R-101$\times$3 & 52.0 ({\color{gray}{5}}) & \multirow{2}{*}{\color{red}{0.3}} & 82.5 ({\color{gray}{5}}) & \multirow{2}{*}{\color{red}{2.1}}& \multirow{2}{*}{{\color{gray}{4.5}}} \\
         R-152$\times$4 & 52.8 ({\color{gray}{4}}) & & 83.4 ({\color{gray}{4}}) & &  \\
         \hline
         ViT-B/16 & 53.2 ({\color{gray}{3}}) & \multirow{3}{*}{\color{red}{0.1}} & 83.7 ({\color{gray}{3}}) & \multirow{3}{*}{\color{red}{1.3}} & \multirow{3}{*}{{\color{gray}{2.0}}}\\
         ViT-L/16 & 53.5 ({\color{gray}{2}}) & & 84.2 ({\color{gray}{2}}) &\\
         Swin-B & 53.8 ({\color{gray}{1}}) & & 86.5 ({\color{gray}{1}}) &\\
         \bottomrule
    \end{tabular}
    \caption{Multi-task learning on scene segmentation and classification. The performance ranks (in a descending order) are marked using color grey. Color green/red denotes the relative improvement/drop over single-task training.}
    \vspace{-1.0em}
    \label{scene_cls_seg}
\end{table*}
\begin{table*}[]
    \centering
    \scriptsize
    \begin{tabular}{c|c|c|c|c|c}
        \toprule
         \multirow{2}{*}{Model} & \multicolumn{2}{c|}{NYU segmentation (mIoU)} & \multicolumn{2}{c|}{WikiArt (acc.)} & Ave. rank $\downarrow$ \\ \cline{2-5}
         & Perf. $\uparrow$ & Ave. {\color{ForestGreen}Imp. $\uparrow$} / {\color{red}Drop $\downarrow$} & Perf. $\uparrow$ & Ave. {\color{ForestGreen}Imp. $\uparrow$} / {\color{red}Drop $\downarrow$} &\\
         \hline
         R-101$\times$3 & 50.9 ({\color{gray}{3}}) & \multirow{2}{*}{\color{red}{3.3}} & 65.7 ({\color{gray}{5}}) &\multirow{2}{*}{\color{red}{0.4}} & \multirow{2}{*}{{\color{gray}{4.3}}} \\
         R-152$\times$4 & 47.8 ({\color{gray}{5}}) & & 66.1 ({\color{gray}{4}}) & & \\
         \hline
         ViT-B/16 & 51.7 ({\color{gray}{2}}) & \multirow{3}{*}{\color{red}{2.3}} & 67.1 ({\color{gray}{3}}) &\multirow{3}{*}{\color{red}{0.5}} & \multirow{3}{*}{{\color{gray}{2.2}}}\\
         ViT-L/16 & 49.2 ({\color{gray}{4}}) & & 67.3 ({\color{gray}{2}}) & \\
         Swin-B & 52.8 ({\color{gray}{1}}) & & 70.8 ({\color{gray}{1}}) & \\
         \bottomrule
    \end{tabular}
    \caption{Multi-task learning on scene segmentation and open-domain classification. Colors grey, green and red denote the performance ranks, relative performance improvement and drop, respectively.}
    \vspace{-1.0em}
    \label{scene_seg_open}
\end{table*}

\subsection{Face recognition}
\noindent \textbf{FG-NET\footnote{\url{https://yanweifu.github.io/FG_NET_data/}}.} FG-NET consists of 1,002 color or gray facial images of more than 50 individuals with large variations in pose, expression and lighting. For each subject, there are more than ten images ranging from age 0 to age 69.\\

\noindent \textbf{Implementation details.} We discuss some specific details about the experiment on FG-Net as follows. For other details, we simply follow the operations on Flower102 and CUB200.
\begin{itemize}
    \item Data split: We randomly choose 5 individuals and 3 individuals to build the test and validation sets, respectively. The remaining images are used for training.
    \item Augmentation strategies: Common augmentation methods like random crop and random horizontal flip are adopted. Besides, we also employ random affine, where the rotation degree is randomly chosen between -10 degrees and 10 degrees and a shear operation (between -12 degrees and 12 degrees) parallel to the x-axis is also applied. The input size is 224$\times$224.
    \item Loss function: We use two loss functions: mean variance loss \cite{pan2018mean} (weight=1) and cross entropy loss (weight=1).
\end{itemize}
\noindent \textbf{Results.} Since ImageNet pre-training does not contain many face images, the performance on face recognition can somewhat reflect the transferring and generalization abilities of representations (like the problem of open-domain classification). As shown in Table \ref{FG-NET}, vision transformers again achieve a higher average rank than ConvNet-based backbones, demonstrating Transformers are more capable of tackling open-domain problems that are more practical and applicable for real-world applications. Another interesting phenomenon is that bigger models like R-152$\times$4 and ViT-L/16 are outperformed by R-101$\times$3 and ViT-B/16, respectively, while Swin-B still takes the first place. Such observations imply that efficient models (with fewer parameters) are more suitable for face recognition tasks as they reduce the risking of overfitting.

\subsection{Linear evaluation protocol}
\label{linear_eval}
In this section, we first perform linear classification using ConvNet- and Transformer-based pre-trained feature representations directly by freezing the backbone and training a supervised linear classification head (i.e., a fully-connected layer followed by softmax) on CUB200, Indoor67 and WikiArt, respectively. In addition, we also conduct unsupervised image retrieval using pre-trained feature representations from each backbone directly.\\

\textbf{Implementation details.} In linear classification, we train the classifier on the global average pooling features for ConvNet-based backbones and Swin-B. All training details are similar to those of fine-tuning. In unsupervised image retrieval, we first resize each input image to 256$\times$256, after which we apply center crop to generate a central patch whose size is 224$\times$224. Next, we forward each central patch to the backbone network and apply L2 normalization to its extracted feature representation. The evaluation metric is mean average precision (mAP).\\

\noindent \textbf{Results.} Table \ref{linear_cls} reports the results of linear classification. Somewhat surprisingly, ConvNet-based backbones achieve a comparable average rank with that of vision transformers. More specifically, it seems that ViT-B/16 and ViT-L/16 lack the ability to provide as transferable representations as they did in fine-tuning. Nonetheless, Swin-B still achieves first places on all 3 datasets, again verifying the effectiveness of feature pyramids. In contrast to the fine-tuning improvements of vision transformers on 3 datasets (presented in Tables \ref{fine-grained}, \ref{scene} and \ref{open}), experimental results in Table \ref{linear_cls} imply that Transformer-based backbones are more advantageous in fine-tuning than ConvNet-based models. From Table \ref{retrieval}, we can see that ConvNet-based backbones have more advantages on the task of unsupervised image retrieval. The underlying reason might be that images in Oxford5 are very similar to those from ImageNet where R-101$\times$3 and R-152$\times$4 produce better classification performance. On the other hand, we notice that Swin-B exhibits much better retrieval results than ViT-B/16 and ViT-L/16. Since Swin-B employs a pyramidal hierarchy scheme to learn visual representations as ConvNets, we think it could be beneficial to learn features for image retrieval in a hierarchical manner using vision transformers.

\section{Multi-task evaluation}
In this section, we evaluate ConvNet- and Transformer-based backbones on 6 multi-task learning problems. Apart from exact experimental results and corresponding performance ranks, we also present relative performance improvement/drop over single-task learning. Note that we do not strictly require that the individual tasks in each multi-task problem are beneficial to each other (i.e., providing improvements over single tasks). Instead, our goal is to investigate which type of models would better handle the problem of multi-task learning, which is quite necessary and applicable in real-world applications. Similar to single-task evaluation, we repeat each experiment for 3 times and report the average results. 

\subsection{Scene segmentation and depth estimation}
In this setting, each image is associated with segmentation labels and depth values, both of which are acquired from NYU Depth V2 dataset.\\

\noindent \textbf{Implementation details.} We include two heads in UPerNet, where the original segmentation head is used for scene segmentation and depth prediction head is responsible for depth estimation. For loss functions, we sum up those used in single-task training with equal weights (weight=1). We resize each image to 384$\times$384 and the training batch size is 8. For other details, we directly follow those used in single-task learning.\\

\noindent \textbf{Results.} From Table \ref{scene_seg_depth}, we can draw a conclusion that scene segmentation and depth estimation are beneficial to each other because the multi-task learning leads to better performance on both tasks over single-task learning. In comparison to ConvNet-based models, vision transformers utilize complementary information hidden in two tasks more effectively, which can be verified by the larger improvements on both tasks (i.e., 0.4 vs. 0.3 on segmentation and 0.007 vs. 0.005 on depth estimation). Not surprisingly, Transformer-based backbones again hold a higher average performance rank than ConvNet-based models.

\subsection{Scene and in-domain classification}
\label{sec_scene_and_in_domain}
We perform multi-task classification on a combination of scene and in-domain classification datasets, where each image comes from either Indoor67 or Caltech101, as shown in Table \ref{scene_and_in_domain}.\\

\noindent \textbf{Implementation details.} Images from Indoor67 and Caltech101 are randomly shuffled to build a mixed dataset. During the training stage, we append two classification heads to the backbone, each with a dropout layer ($p$=0.2). To distinguish images from Indoor67 and those from Caltech101, we add a pre-classification head whose output size is 1. We apply cross entropy loss to train both classification and pre-classification heads. For inference, we take two steps to make predictions. For each test input, we first decide whether it belongs to Indoor67 or Caltech101 according to the output of the pre-classification head (after the sigmoid function). Then, we compute mean accuracy on each dataset, respectively. For other training and inference details, we follow the operations on Flower102 and CUB200.\\

\noindent \textbf{Results.} From Table \ref{scene_and_in_domain}, we can see that Transformer-based backbones still maintain observable advantages (i.e., higher performance rank) over ConvNets in the problem of scene and in-domain classification. Specifically, Swin-B again takes the first places on both Indoor67 and Caltech101, demonstrating its capability to deal with scene and in-domain classification, simultaneously. It is not surprising to find that the results on either dataset are slightly lower than those trained on each dataset solely, showing that Indoor67 and Caltech101 are not beneficial to each other. Besides, we find that Transformer-based backbones suffer from smaller performance drops on Indoor67 and perform on-par with single-task classification on Caltech101 using multi-task learning. In comparison, ConvNet-based models suffer from larger performance drops on both datasets.

\subsection{Open-domain and in-domain classification}
In this part, we replace the scene dataset (i.e., Indoor67) with WikiArt to investigate multi-task learning on a combination of open-domain and in-domain classification problems. Experimental results are presented in Table \ref{open_and_in}.\\

\noindent \textbf{Implementation details.} We directly refer to the details on scene and in-domain classification (refer to Sec. \ref{sec_scene_and_in_domain}).\\

\noindent \textbf{Results.} From Table \ref{open_and_in}, we can see that Transformer-based backbones outperform ConvNet-based models by large margins in multi-task learning for both art style and generic object recognition. It is understandable that conducting multi-task training using both Caltech101 and WikiArt does not boost the performance over single-task training as the involved images come from two dramatically different domains and thus require different representations towards the goal of recognition. Nonetheless, Transformer-based backbones can still greatly reduce the performance drop by nearly 5 percents on Caltech101, compared to ConvNets, again verifying the ability of vision transformers in dealing with two non-related tasks.

\subsection{Scene segmentation with different classification problems}
We investigate the possibility of incorporating scene segmentation into scene classification and open-domain classification, respectively. Experimental results are given in Tables \ref{scene_cls_seg} and \ref{scene_seg_open}.\\

\noindent \textbf{Implementation details.} For network architecture, we use UPerNet to carry out segmentation. To perform classification tasks at the same time, we replace the fully-connected layers for ImageNet pre-training with classification heads for different classification tasks. Specifically, for ConvNet-based backbones and Swin-B, we add classification heads on global average pooling features in their last layers (i.e., right before the upsampling layer in the bottleneck of UPerNet). The initial learning rate is 1e-4, and we follow the operations in single-task scene segmentation to decrease learning rate and conduct warm-up. The number of training iterations is 20,000. The input size is 384$\times$384. We employ random crop and random horizontal flip as default augmentation strategies.\\

\noindent \textbf{Results.} It is observable that Transformer-based backbones maintain consistent and significant advantages (i.e., higher average performance ranks) over ConvNets in all combinations. Besides, we can still find that it is hard to conduct multi-task learning on top of segmentation and different classification problems even when the incorporated classification problem is closely related to scenes (i.e., Indoor67). Nonetheless, vision transformers are more resistant to tasks that are hard to combine than ConvNets, which implies that Transformer may be a better choice than ConvNet when dealing with unknown multi-task problems.

\section{Conclusion}
We found that Transformer-based backbones provide more transferable representations than ConvNets for fine-tuning, especially when the downstream tasks come from domains very different from ImageNet, which is used for pre-training. Meanwhile, vision transformers are more robust in multi-task learning, where they achieve larger improvements and suffer from smaller performance losses. On the other hand, we observe that ConvNets still have slight advantages on in-domain classification and unsupervised image retrieval. In our future work, we will include more datasets in these two types of problems to obtain more comprehensive results and find out the underlying reasons.

{\small
\bibliographystyle{ieee_fullname}
\bibliography{egbib}

\begin{thebibliography}{10}\itemsep=-1pt

\bibitem{bahdanau2014neural}
Dzmitry Bahdanau, Kyunghyun Cho, and Yoshua Bengio.
\newblock Neural machine translation by jointly learning to align and
  translate.
\newblock {\em arXiv preprint arXiv:1409.0473}, 2014.

\bibitem{cohen2020covid}
Joseph~Paul Cohen, Paul Morrison, Lan Dao, Karsten Roth, Tim~Q Duong, and
  Marzyeh Ghassemi.
\newblock Covid-19 image data collection: Prospective predictions are the
  future.
\newblock {\em arXiv preprint arXiv:2006.11988}, 2020.

\bibitem{deng2009imagenet}
Jia Deng, Wei Dong, Richard Socher, Li-Jia Li, Kai Li, and Li Fei-Fei.
\newblock Imagenet: A large-scale hierarchical image database.
\newblock In {\em Proceedings of the IEEE Conference on Computer Vision and
  Pattern Recognition}, pages 248--255, 2009.

\bibitem{dosovitskiy2020image}
Alexey Dosovitskiy, Lucas Beyer, Alexander Kolesnikov, Dirk Weissenborn,
  Xiaohua Zhai, Thomas Unterthiner, Mostafa Dehghani, Matthias Minderer, Georg
  Heigold, Sylvain Gelly, et~al.
\newblock An image is worth 16x16 words: Transformers for image recognition at
  scale.
\newblock {\em arXiv preprint arXiv:2010.11929}, 2020.

\bibitem{fei2004learning}
Li Fei-Fei, Rob Fergus, and Pietro Perona.
\newblock Learning generative visual models from few training examples: An
  incremental bayesian approach tested on 101 object categories.
\newblock In {\em Proceedings of the IEEE Conference on Computer Vision and
  Pattern Recognition Workshops}, pages 178--178. IEEE, 2004.

\bibitem{he2016deep}
Kaiming He, Xiangyu Zhang, Shaoqing Ren, and Jian Sun.
\newblock Deep residual learning for image recognition.
\newblock In {\em Proceedings of the IEEE Conference on Computer Vision and
  Pattern Recognition}, pages 770--778, 2016.

\bibitem{kolesnikov2020big}
Alexander Kolesnikov, Lucas Beyer, Xiaohua Zhai, Joan Puigcerver, Jessica Yung,
  Sylvain Gelly, and Neil Houlsby.
\newblock Big transfer (bit): General visual representation learning.
\newblock In {\em Proceedings of the European Conference on Computer Vision
  (ECCV)}, pages 491--507. Springer, 2020.

\bibitem{kornblith2019better}
Simon Kornblith, Jonathon Shlens, and Quoc~V Le.
\newblock Do better imagenet models transfer better?
\newblock In {\em Proceedings of the IEEE Conference on Computer Vision and
  Pattern Recognition}, pages 2661--2671, 2019.

\bibitem{krizhevsky2012imagenet}
Alex Krizhevsky, Ilya Sutskever, and Geoffrey~E Hinton.
\newblock Imagenet classification with deep convolutional neural networks.
\newblock {\em Advances in Neural Information Processing Systems},
  25:1097--1105, 2012.

\bibitem{li2018megadepth}
Zhengqi Li and Noah Snavely.
\newblock Megadepth: Learning single-view depth prediction from internet
  photos.
\newblock In {\em Proceedings of the IEEE Conference on Computer Vision and
  Pattern Recognition}, pages 2041--2050, 2018.

\bibitem{liu2021swin}
Ze Liu, Yutong Lin, Yue Cao, Han Hu, Yixuan Wei, Zheng Zhang, Stephen Lin, and
  Baining Guo.
\newblock Swin transformer: Hierarchical vision transformer using shifted
  windows.
\newblock {\em arXiv preprint arXiv:2103.14030}, 2021.

\bibitem{loshchilov2017decoupled}
Ilya Loshchilov and Frank Hutter.
\newblock Decoupled weight decay regularization.
\newblock {\em arXiv preprint arXiv:1711.05101}, 2017.

\bibitem{pan2018mean}
Hongyu Pan, Hu Han, Shiguang Shan, and Xilin Chen.
\newblock Mean-variance loss for deep age estimation from a face.
\newblock In {\em Proceedings of the IEEE Conference on Computer Vision and
  Pattern Recognition}, pages 5285--5294, 2018.

\bibitem{paszke2019pytorch}
Adam Paszke, Sam Gross, Francisco Massa, Adam Lerer, James Bradbury, Gregory
  Chanan, Trevor Killeen, Zeming Lin, Natalia Gimelshein, Luca Antiga, et~al.
\newblock Pytorch: An imperative style, high-performance deep learning library.
\newblock {\em Advances in Neural Information Processing Systems},
  32:8026--8037, 2019.

\bibitem{quattoni2009recognizing}
Ariadna Quattoni and Antonio Torralba.
\newblock Recognizing indoor scenes.
\newblock In {\em 2009 IEEE Conference on Computer Vision and Pattern
  Recognition}, pages 413--420. IEEE, 2009.

\bibitem{ranftl2021vision}
Ren{\'e} Ranftl, Alexey Bochkovskiy, and Vladlen Koltun.
\newblock Vision transformers for dense prediction.
\newblock {\em arXiv preprint arXiv:2103.13413}, 2021.

\bibitem{sharif2014cnn}
Ali Sharif~Razavian, Hossein Azizpour, Josephine Sullivan, and Stefan Carlsson.
\newblock Cnn features off-the-shelf: an astounding baseline for recognition.
\newblock In {\em Proceedings of the IEEE Conference on Computer Vision and
  Pattern Recognition Workshops}, pages 806--813, 2014.

\bibitem{silberman2012indoor}
Nathan Silberman, Derek Hoiem, Pushmeet Kohli, and Rob Fergus.
\newblock Indoor segmentation and support inference from rgbd images.
\newblock In {\em Proceedings of the European Conference on Computer Vision
  (ECCV)}, pages 746--760. Springer, 2012.

\bibitem{simonyan2014very}
Karen Simonyan and Andrew Zisserman.
\newblock Very deep convolutional networks for large-scale image recognition.
\newblock {\em arXiv preprint arXiv:1409.1556}, 2014.

\bibitem{szegedy2015going}
Christian Szegedy, Wei Liu, Yangqing Jia, Pierre Sermanet, Scott Reed, Dragomir
  Anguelov, Dumitru Erhan, Vincent Vanhoucke, and Andrew Rabinovich.
\newblock Going deeper with convolutions.
\newblock In {\em Proceedings of the IEEE Conference on Computer Vision and
  Pattern Recognition}, pages 1--9, 2015.

\bibitem{tan2016ceci}
Wei~Ren Tan, Chee~Seng Chan, Hern{\'a}n~E Aguirre, and Kiyoshi Tanaka.
\newblock Ceci n'est pas une pipe: A deep convolutional network for fine-art
  paintings classification.
\newblock In {\em 2016 IEEE international conference on image processing
  (ICIP)}, pages 3703--3707. IEEE, 2016.

\bibitem{vaswani2017attention}
Ashish Vaswani, Noam Shazeer, Niki Parmar, Jakob Uszkoreit, Llion Jones,
  Aidan~N Gomez, {\L}ukasz Kaiser, and Illia Polosukhin.
\newblock Attention is all you need.
\newblock In {\em Advances in Neural Information Processing Systems}, pages
  5998--6008, 2017.

\bibitem{wah2011caltech}
Catherine Wah, Steve Branson, Peter Welinder, Pietro Perona, and Serge
  Belongie.
\newblock The caltech-ucsd birds-200-2011 dataset.
\newblock 2011.

\bibitem{welinder2010caltech}
Peter Welinder, Steve Branson, Takeshi Mita, Catherine Wah, Florian Schroff,
  Serge Belongie, and Pietro Perona.
\newblock Caltech-ucsd birds 200.
\newblock 2010.

\bibitem{xiao2018unified}
Tete Xiao, Yingcheng Liu, Bolei Zhou, Yuning Jiang, and Jian Sun.
\newblock Unified perceptual parsing for scene understanding.
\newblock In {\em Proceedings of the European Conference on Computer Vision
  (ECCV)}, pages 418--434, 2018.

\bibitem{xie2017aggregated}
Saining Xie, Ross Girshick, Piotr Doll{\'a}r, Zhuowen Tu, and Kaiming He.
\newblock Aggregated residual transformations for deep neural networks.
\newblock In {\em Proceedings of the IEEE Conference on Computer Vision and
  Pattern Recognition}, pages 1492--1500, 2017.

\bibitem{yosinski2014transferable}
Jason Yosinski, Jeff Clune, Yoshua Bengio, and Hod Lipson.
\newblock How transferable are features in deep neural networks?
\newblock {\em Advances in Neural Information Processing Systems},
  27:3320--3328, 2014.

\bibitem{zoph2016neural}
Barret Zoph and Quoc~V Le.
\newblock Neural architecture search with reinforcement learning.
\newblock {\em arXiv preprint arXiv:1611.01578}, 2016.

\end{thebibliography}
}

\end{document}